# Definition and Complexity of Some Basic Metareasoning Problems[*]

**Vincent Conitzer** and **Tuomas Sandholm**
Carnegie Mellon University
Computer Science Department
5000 Forbes Avenue
Pittsburgh, PA 15213, USA
{conitzer,sandholm}@cs.cmu.edu

## Abstract

In most real-world settings, due to limited time or other resources, an agent cannot perform all potentially useful deliberation and information gathering actions. This leads to the metareasoning problem of selecting such actions. Decision-theoretic methods for metareasoning have been studied in AI, but there are few theoretical results on the complexity of metareasoning. We derive hardness results for three settings which most real metareasoning systems would have to encompass as special cases. In the first, the agent has to decide how to allocate its deliberation time across anytime algorithms running on different problem instances. We show this to be $\mathcal{NP}$-complete. In the second, the agent has to (dynamically) allocate its deliberation or information gathering resources across multiple actions that it has to choose among. We show this to be $\mathcal{NP}$-hard even when evaluating each individual action is extremely simple. In the third, the agent has to (dynamically) choose a limited number of deliberation or information gathering actions to disambiguate the state of the world. We show that this is $\mathcal{NP}$-hard under a natural restriction, and $\mathcal{PSPACE}$-hard in general.

## 1 Introduction

In most real-world settings, due to limited time, an agent cannot perform all potentially useful deliberation actions. As a result it will generally be unable to act rationally in the world. This phenomenon, known as bounded rationality, has been a long-standing research topic (e.g., [3, 17]). Most of that research has been *descriptive*: the goal has been to characterize how agents—in particular, humans—deal with this constraint. Another strand of bounded rationality research has the *normative (prescriptive)* goal of characterizing how agents *should* deal with this constraint. This is particularly important when building artificial agents.

Characterizing how an agent should deal with bounded rationality entails determining how the agent should deliberate.

[*]The material in this paper is based upon work supported by the National Science Foundation under CAREER Award IRI-9703122, Grant IIS-9800994, ITR IIS-0081246, and ITR IIS-0121678.

Because limited time (or other resources) prevent the agent from performing all potentially useful deliberation (or information gathering) actions, it has to select among such actions. Reasoning about which deliberation actions to take is called *metareasoning*. Decision theory [7, 10] provides a normative basis for metareasoning under uncertainty, and decision-theoretic deliberation control has been widely studied in AI (e.g., [2, 4–6, 8, 9, 12–15, 18–20]).

However, the approach of using metareasoning to control reasoning is impractical if the metareasoning problem itself is prohibitively complex. While this issue is widely acknowledged (e.g., [8, 12–14]), there are few theoretical results on the complexity of metareasoning.

We derive hardness results for three central metareasoning problems. In the first (Section 2), the agent has to decide how to allocate its deliberation time across anytime algorithms running on different problem instances. We show this to be $\mathcal{NP}$-complete. In the second metareasoning problem (Section 3), the agent has to (dynamically) allocate its deliberation or information gathering resources across multiple actions that it has to choose among. We show this to be $\mathcal{NP}$-hard even when evaluating each individual action is extremely simple. In the third metareasoning problem (Section 4), the agent has to (dynamically) choose a limited number of deliberation or information gathering actions to disambiguate the state of the world. We show that this is $\mathcal{NP}$-hard under a natural restriction, and $\mathcal{PSPACE}$-hard in general.

These results have general applicability in that most metareasoning systems must somehow deal with one or more of these problems (in addition to dealing with other issues). We also believe that these results give a good basic overview of the space of high-complexity issues in metareasoning.

## 2 Allocating anytime algorithm time across problems

In this section we study the setting where an agent has to allocate its deliberation time across different problems—each of which the agent can solve using an anytime algorithm. We show that this is hard even if the agent can perfectly predict the performance of the anytime algorithms.

### 2.1 Motivating example

Consider a newspaper company that has, by midnight, received the next day's orders from newspaper stands in the 3

cities where the newspaper is read. The company owns a fleet of delivery trucks in each of the cities. Each fleet needs its vehicle routing solution by 5am. The company has a default routing solution for each fleet, but can save costs by improving (tailoring to the day's particular orders) the routing solution of any individual fleet using an anytime algorithm. In this setting, the "solution quality" that the anytime algorithm provides on a fleet's problem instance is the amount of savings compared to the default routing solution.

We assume that the company can perfectly predict the savings made on a given fleet's problem instance as a function of deliberation time spent on it (we will prove hardness of metareasoning even in this deterministic variant). Such functions are called *(deterministic) performance profiles* [2, 6, 8, 9, 20]. Each fleet's problem instance has its own performance profile.[1] Suppose the performance profiles are as shown in Fig. 1.

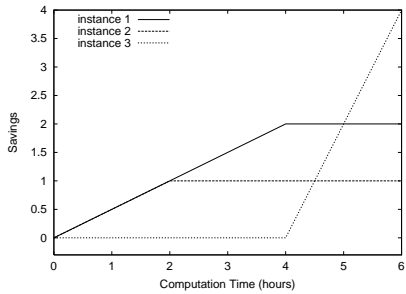

Figure 1: Performance profiles for the routing problems.

Then the maximum savings we can obtain with 5 hours of deliberation time is 2.5, for instance by spending 3 hours on instance 1 and 2 on instance 2. On the other hand, if we had until 6am to deliberate (6 hours), we could obtain a savings of 4 by spending 6 hours on instance 3.

## 2.2 Definitions and results

We now define the metareasoning problem of allocating deliberation across problems according to performance profiles.

**Definition 1 (PERFORMANCE-PROFILES)** *We are given a list of performance profiles $(f_1, f_2, \ldots, f_m)$ (where each $f_i$ is a nondecreasing function of deliberation time, mapping to nonnegative real numbers), a number of deliberation steps $N$, and a target value $K$. We are asked whether we can distribute the deliberation steps across the problem instances to get a total performance of at least $K$; that is, whether there exists a vector $(N_1, N_2, \ldots, N_m)$ with $\sum_{1 \leq i \leq m} N_i \leq N$ and $\sum_{1 \leq i \leq m} f_i(N_i) \geq K$.*

A reasonable approach to representing the performance profiles is to use piecewise linear performance profiles. They can model any performance profile arbitrarily closely, and have been used in the resource-bounded reasoning literature to characterize the performance of anytime algorithms (e.g. [2]). We now show that the metareasoning problem is $\mathcal{NP}$-complete even under this restriction. We will reduce from the KNAPSACK problem.[2]

**Definition 2 (KNAPSACK)** *We are given a set $S$ of $m$ pairs of positive integers $(c_i, v_i)$, a constraint $C > 0$ and a target value $V > 0$. We are asked whether there exists a set $I \subseteq S$ such that $\sum_{j \in I} C_j \leq C$ and $\sum_{j \in I} v_j \geq V$.*

**Theorem 1** *PERFORMANCE-PROFILES is $\mathcal{NP}$-complete even if each performance profile is continuous and piecewise linear.*[3]

**Proof**: The problem is in $\mathcal{NP}$ because we can nondeterministically generate the $N_i$ in polynomial time (since we do not need to bother trying numbers greater than $N$), and given the $N_i$, we can verify if the target value is reached in polynomial time. To show $\mathcal{NP}$-hardness, we reduce an arbitrary KNAPSACK instance to the following PERFORMANCE-PROFILES instance. Let there be $m$ performance profiles, given by $f_i(n) = 0$ for $n \leq c_i$, $f_i(n) = n - c_i$ for $c_i < n \leq c_i + v_i$, and $f_i(n) = v_i$ for $n > c_i + v_i$. Let $N = C + V$ and let $K = V$. We claim the two problem instances are equivalent. First suppose there is a solution to the KNAPSACK instance, that is, a set $I \subseteq S$ such that $\sum_{j \in I} C_j \leq C$ and $\sum_{j \in I} v_j \geq V$. Then set the $N_i$ as follows: $N_i = 0$ for $i \notin I$, and $N_i = c_i + \frac{V}{\sum_{j \in I} v_j} v_i$ for $i \in I$. Then, $\sum_{1 \leq i \leq m} N_i = \sum_{i \in I} N_i = \sum_{i \in I} (c_i + \frac{V}{\sum_{j \in I} v_j} v_i) = \sum_{i \in I} c_i + \frac{V}{\sum_{j \in I} v_j} \sum_{i \in I} v_i = \sum_{i \in I} c_i + V \leq C + V = N$ since $\sum_{j \in I} C_j \leq C$. Also, since $\sum_{j \in I} v_j \geq V$, it follows that for every $i \in I$, we have $c_i \leq N_i \leq c_i + v_i$, and hence $\sum_{1 \leq i \leq m} f_i(N_i) = \sum_{i \in I} f_i(N_i) = \sum_{i \in I} (\frac{V}{\sum_{j \in I} v_j} v_i) =$ $\frac{V}{\sum_{j \in I} v_j} \sum_{i \in I} v_i = V = K$. So we have found a solution to the PERFORMANCE-PROFILES instance. On the other hand, suppose there is a solution to the PERFORMANCE-PROFILES instance, that is, a vector $(N_1, N_2, \ldots, N_m)$ with $\sum_{1 \leq i \leq m} N_i \leq N$ and $\sum_{1 \leq i \leq m} f_i(N_i) \geq K$. Since spending more than $c_i + v_i$ deliberation steps on profile $i$ is useless, we may assume that $N_i \leq c_i + v_i$ for all $i$. We now claim that $I = \{i : N_i \geq c_i\}$ is a solution to the KNAPSACK instance. First, using the fact that $f_j(N_j) = 0$ for all $j \notin I$, we have $\sum_{i \in I} v_i \geq \sum_{i \in I} f_i(N_i) = \sum_{1 \leq i \leq m} f_i(N_i) \geq K = V$. Also, $\sum_{i \in I} c_i = \sum_{i \in I} N_i - \sum_{i \in I}(N_i - c_i) = \sum_{i \in I} N_i - \sum_{i \in I} f_i(N_i) \leq \sum_{1 \leq i \leq m} N_i - \sum_{1 \leq i \leq m} f_i(N_i) \leq N - K = C + V - V = C$. So we have found a solution to the KNAPSACK instance. ∎

---

[1]Because the anytime algorithm's performance differs across instances, each instance has its own performance profile (in the setting of deterministic performance profiles). In reality, an anytime algorithm's performance on an instance cannot be predicted perfectly. Rather, usually statistical performance profiles are kept that aggregate across instances. In that light one might question the assumption that different instances have different performance profiles. However, sophisticated deliberation control systems can condition the performance prediction on features of the instance—and this is necessary if the deliberation control is to be fully normative. (Research has already been conducted on conditioning performance profiles on instance features [8, 9, 15] or results of deliberation on the instance so far [4, 8, 9, 15, 18–20].)

[2]This only demonstrates *weak* NP-completeness, as KNAPSACK is weakly NP-complete; thus, perhaps pseudopolynomial time algorithms exist.

[3]If one additionally assumes that each performance profile is concave, then the metareasoning problem is solvable in polynomial time [2]. While returns to deliberation indeed tend to be diminishing, usually this is not the case throughout the performance profile. Algorithms often have a setup phase in the beginning during which there is no improvement. Also, iterative improvement algorithms can switch to using different local search operators once progress has ceased using one operator (for example, once 2-swap has reached a local optimum in TSP, one can switch to 3-swap and obtain gains from deliberation again) [16].

The PERFORMANCE-PROFILES problem occurs naturally as a subproblem within many metareasoning problems, and thus its complexity leads to significant difficulties for metareasoning. This is the case even under the (unrealistic) assumption of perfect predictability of the efficacy of deliberation. On the other hand, in the remaining two metareasoning problems that we analyze, the complexity stems from uncertainty about the results that deliberation will provide.

## 3 Dynamically allocating evaluation effort across options (actions)

In this section we study the setting where an agent is faced with multiple options (actions) from which it eventually has to choose one. The agent can use deliberation (or information gathering) to evaluate each action. Given limited time, it has to decide which ones to evaluate. We show that this is hard even in very restricted cases.

### 3.1 Motivating example

Consider an autonomous robot looking for precious metals. It can choose between three sites for digging (it can dig at most one site). At site $A$ it may find gold; at site $B$, silver; at site $C$, copper. If the robot chooses not to dig anywhere, it gets utility 1 (for saving digging costs). If the robot chooses to dig somewhere, the utility of finding nothing is 0; finding gold, 5; finding silver, 3; finding copper, 2. The prior probability of there being gold at site $A$ is $\frac{1}{8}$, that of finding silver at site $B$ is $\frac{1}{2}$, and that of finding copper at site $C$ is $\frac{1}{2}$.

In general, the robot could perform deliberation or information gathering actions to evaluate the alternative (digging) actions. The metareasoning problem would be the same for both, so for simplicity of exposition, we will focus on information gathering only. Specifically, the robot can perform tests to better evaluate the likelihood of there being a precious metal at each site, but it has only limited time for such tests. The tests are the following: (1) Test for gold at $A$. If there is gold, the test will be positive with probability $\frac{14}{15}$; if there is no gold, the test will be positive with probability $\frac{1}{15}$. This test takes 2 units of time. (2) Test for silver at $B$. If there is silver, the test will be positive with probability 1; if there is no silver, the test will be positive with probability 0. This test takes 3 units of time. (3) Test for copper at $C$. If there is copper, the test will be positive with probability 1; if there is no copper, the test will be positive with probability 0. This test takes 2 units of time.

Given the probabilities of the tests turning out positive under various circumstances, one can use Bayes' rule to compute the expected utility of each digging option given any (lack of) test result. For instance, letting $a$ be the event that there is gold at $A$, and $t_A$ be the event that the test at $A$ is positive, we observe that $P(t_A) = P(t_A|a)P(a) + P(t_A|-a)P(-a) = \frac{14}{15}\frac{1}{8} + \frac{1}{15}\frac{7}{8} = \frac{7}{40}$. Then, the expected utility of digging at $A$ given that the test at $A$ was positive is $5\ P(a|t_A)$, where $P(a|t_A) = \frac{P(t_A|a)P(a)}{P(t_A)} = \frac{\frac{14}{15}\frac{1}{8}}{\frac{7}{40}} = \frac{2}{3}$, so the expected utility is $\frac{10}{3}$. Doing a similar analysis everywhere, we can represent the problem by trees shown in Fig. 2. In these trees, be-ing at the root represents not having done a test yet, whereas being at a left (right) leaf represents the test having turned out positive (negative); the value at each node is the expected value of digging at this site given the information corresponding to that node. The values on the edges are the probabilities of the test turning out positive or negative. We can subsequently use these trees for analyzing how we should gather information. For instance, if we have 5 units of time, the optimal information gathering policy is to test at $B$ first; if the result is positive, test at $A$; otherwise test at $C$. (We omit the proof because of space constraint.)

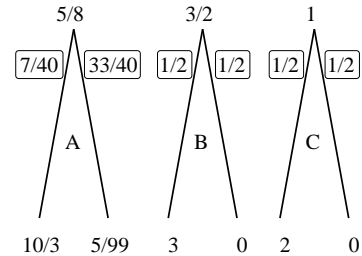

Figure 2: Tree representation of the action evaluation instance.

### 3.2 Definitions

In the example, there were four *actions* that we could evaluate: digging for a precious metal at one of three locations, or not digging at all. Given the results of all the tests that we might undertake on a given action, executing it has some expected value. If, on the other hand, we do not (yet) know all the results of these tests, we can still associate an expected value with the action by taking an additional expectation over the outcomes of the tests. In what follows, we will drop the word "expected" in its former meaning (that is, when talking about the expected value given the outcomes of all the tests), because the probabilistic process regarding this expectation has no relevance to how the agent should choose to test. Hence, all expectations are over the outcomes of the tests.

While we have presented this as a model for information gathering planning, we can use this as a model for planning

(computational) deliberation over multiple actions as well. In this case, we regard the tests as computational steps that the agent can take toward evaluating an action.[4]

To proceed, we need a formal model of how evaluation effort (information gathering or deliberation) invested on a given action changes the agent's beliefs about that action. For this model, we generalize the above example to the case where we can take multiple evaluation steps on a certain action (although we will later show hardness even when we can take at most one evaluation step per action).

**Definition 3** *An* action evaluation tree *is a tree with*

- *A root $r$, representing the start of the evaluation;*
- *For each nonleaf node $w$, a cost $k_w$ for investing another step of evaluation effort at this point;*
- *For each edge $e$ between parent node $p$ and child node $c$, a probability $p_e = p_{(p,c)}$ of transitioning from $p$ to $c$ upon taking a step of evaluation effort at $p$;*
- *For each leaf node $l$, a value $u_l$.*

According to this definition, at each point in the evaluation of a single action, the agent's only choice is *whether* to invest further evaluation effort, but not *how* to continue the evaluation. This is a reasonable model when the agent does evaluation through deliberation and has one algorithm at its disposal. However, in general the agent may have different information gathering actions to choose from at a given point in the evaluation, or may be able to choose from among several deliberation actions (e.g., via search control [1, 14]). In Section 4, we will discuss how being able to choose between tests may introduce drastic complexity even when evaluating a single thing. In this section, however, our focus is on the complexities introduced by having to choose between different actions on which to invest evaluation effort next.

The agent can determine its expected value of an action, given its evaluation so far, using the subtree of the action evaluation tree that is rooted at the node where evaluation has brought us so far. This value can be determined in that subtree by propagating upwards from the leafs: for parent $p$ with a set of children $C$, we have $u_p = \sum_{c \in C} (p_{(p,c)} u_c)$.

We now present the metareasoning problem. In general, the agent could use an *online* evaluation control policy where the choices of how to invest future evaluation effort can depend on evaluation results obtained so far. However, to avoid trivial complexity issues introduced by the fact that such a *contingency* strategy for evaluation can be exponential in size, we merely ask what action the agent should invest its first evaluation step on.

**Definition 4 (ACTION-EVALUATION)** *We are given $l$ action evaluation trees, indexed 1 through $l$, corresponding to $l$ different actions. (The transition processes of the trees are independent.) Additionally, we are given an integer $N$. We are asked whether, among the online evaluation control policies that spend at most $N$ units of effort, there exists one that takes its first evaluation step on action 1, and gives maximal expected utility among online evaluation control policies that spend at most $N$ units of effort. (If at the end of the deliberation process, we are at node $w_i$ for tree $i$, then our utility is $\max_{1 \leq i \leq m} u_{w_i}$, because we will choose the action with the highest expected value.)*

### 3.3 Results

We now show that even a severely restricted version of this problem is $\mathcal{NP}$-hard.[5]

**Theorem 2** *ACTION-EVALUATION is $\mathcal{NP}$-hard, even when all trees have depth either 0 or 1, branching factor 2, and all leaf values are -1, 0, or 1.*

**Proof**: If action evaluation tree $j$ has depth 1 and branching factor 2, we represent it by $(p_1^j, p_2^j, u_1^j, u_2^j, k^j)$ where $p_i^j$ is the transition probability to leaf $i$, $u_i^j$ is the value at leaf $i$, and $k^j$ is the cost of taking the (only) step of evaluation. We reduce an arbitrary KNAPSACK instance to the following ACTION-EVALUATION instance. Let $l = m + 3$, let $\delta = \frac{1}{16m(\sum_{1 \leq i \leq m} v_i)^2}$, and let $\epsilon = 2\delta \sum_{1 \leq i \leq m} v_i = \frac{1}{8m \sum_{1 \leq i \leq m} v_i}$.
We set tree 1 $(p_1^1, p_2^1, u_1^1, u_2^1, k^1) = (\epsilon, 1 - \epsilon, 1, -1, 1)$, and tree 2 $(p_1^2, p_2^2, u_1^2, u_2^2, k^1) = ((1-\epsilon)^m(\epsilon + \delta V), 1 - (1-\epsilon)^m(\epsilon+\delta V), 1, -1, C+1)$[6]. Tree 3 has depth 0 and a value of 0. Finally, for each pair $(c_i, v_i)$ in the KNAPSACK instance, there is an evaluation tree $(p_1^{i+3}, p_2^{i+3}, u_1^{i+3}, u_2^{i+3}, k^{i+3}) = (\delta v_i, 1 - \delta v_i, 1, -1, c_i)$. We set $N = C + 1$. We claim the instances are equivalent. First we make some observations about the constructed ACTION-EVALUATION instance. First, once we determine the value of a action to be 1, choosing this action is certainly optimal regardless of the rest of the deliberation process. Second, if at the end of the deliberation process we have not discovered the value of any action to be 1, then for any of the trees of depth 1, either we have discovered the corresponding action's value to be -1, or we have done no deliberation on it at all. In the latter case, the expected value of the action is always below 0 ($\delta$ is carefully set to achieve this). Hence, we will pick action 3 for value 0. It follows that an optimal deliberation policy is one that maximizes the probability of discovering that a action has value 1. Now, consider the *test set* of a policy, which is the set of actions that the policy would evaluate if no action turned out to have value 1. Then, the probability of discovering that a action has value 1 is simply equal to the probability that at least one of the actions in this set has value 1. So, in this case, the quality of a policy is determined by its test set. Now we observe that any optimal action is either the one that only evaluates action 2 (and then runs out of deliberation time), or one that has action 1 in its test set. (For consider any other policy; since evaluating action 1 has minimal cost, and gives strictly higher probability of discovering a action with value 1 than evaluating on any other action besides 2, simply replacing any other action in the test set with action 1 is possible and improves the policy.) Now suppose there is a solution to the KNAPSACK instance, that is, a set $I \subseteq S$ such that $\sum_{i \in I} C_i \leq C$ and $\sum_{i \in I} v_i \geq V$. Then we can construct a policy which has as test set $J = \{1\} \bigcup \{j : j - 3 \in I\}$. (Evaluating all these actions costs at most $C + 1$ deliberation units.) The probability of at least one of these actions having value 1 is at least the probability that exactly one of them has value 1, which is $\sum_{j \in J} p_1^j \prod_{k \in J, k \neq j} p_2^k \geq \sum_{j \in J} p_1^j (1-\epsilon)^m = (1-\epsilon)^m (\epsilon + \sum_{i \in I} \delta v_i) \geq (1 - \epsilon)^m (\epsilon + \delta V) = p_1^2$. Using our previous observation we can conclude that there is an optimal action that has action 1 in its test set, and since the order in which we evaluate actions in the test set does not matter, there is an optimal policy which evaluates action 1 first. On the other hand, suppose there is no solution to the KNAPSACK instance. Consider a policy which has 1 in its test set, that is, the test set can be expressed as $J = \{1\} \bigcup \{j : j - 3 \in I\}$ for some set $I$. Then we must have $\sum_{i \in I} C_i \leq C$, and since there is no solution to the KNAPSACK instance, it follows that $\sum_{i \in I} v_i \leq V - 1$. But the probability that at least one of the actions in the test set has value 1 is at most $\sum_{j \in J} p_1^j = \epsilon + \sum_{i \in I} \delta v_i \leq \epsilon + \delta(V - 1) = \epsilon + \delta V - \delta$. On the other hand, $p_1^2 = (1 - \epsilon)^m (\epsilon + \delta V) \geq (1 - m\epsilon)(\epsilon + \delta V) \geq \epsilon + \delta V - 2m\epsilon^2$. If we now observe that $2m\epsilon^2 = \frac{1}{32m(\sum_{1 \leq i \leq m} v_i)^2} < \frac{1}{16m(\sum_{1 \leq i \leq m} v_i)^2} = \delta$, it follows that the policy of just evaluating action 2 is strictly better. So, there is no optimal policy which evaluates action 1 first. ∎

---

[4]For this to be a useful model, it is necessary that updating beliefs about the value of an action (after taking a deliberation step) is computationally easy relative to the evaluation problem itself.

[5]ACTION-EVALUATION is trivial for $l = 1$: the answer is "yes" if it is possible to take a step of evaluation. The same is true if there is no uncertainty with regard to the value of any action; in that case any evaluation is irrelevant.

[6]Note that using $m$ in the exponent does not make the reduction exponential in size, because the length of the binary representation of numbers with $m$ in the exponent is linear in $m$.

We have no proof that the general problem is in $\mathcal{NP}$. It is an interesting open question whether stronger hardness results can be obtained for it. For instance, perhaps the general problem is $\mathcal{PSPACE}$-complete.

# 4 Dynamically choosing how to disambiguate state

We now move to the setting where the agent has only one thing to evaluate, but can choose the order of deliberation (or information gathering) actions for doing so. In other words, the agent has to decide how to disambiguate its state. We show that this is hard. (We consider this to be the most significant result in the paper.)

## 4.1 Motivating example

Consider an autonomous robot that has discovered it is on the edge of the floor; there is a gap in front of it. It knows this gap can only be one of three things: a staircase ($S$), a hole ($H$), or a canyon ($C$) (assume a uniform prior distribution over these). The robot would like to continue its exploration beyond the gap. There are three courses of physical action available to the robot: attempt a descent down a staircase, attempt to jump over a hole, or simply walk away. If the gap turns out to be a staircase and the robot descends down it, this gives utility 2. If it turns out to be a hole and the robot jumps over it, this gives utility 1 (discovering new floors is more interesting). If the robot walks away, this gives utility 0 no matter what the gap was. Unfortunately, attempting to jump over a staircase or canyon, or trying to descend into a hole or canyon, has the disastrous consequence of destroying the robot (utility $-\infty$). It follows that if the agent cannot determine with certainty what the gap is, it should walk away.

In order to determine the nature of the gap, the robot can conduct various tests (or *queries*). The tests can determine the answers to the following questions: (1) Am I inside a building? A *yes* answer is consistent only with $S$; a *no* answer is consistent with $S, H, C$. (2) If I drop a small item into the gap, do I hear it hit the ground? A *yes* answer is consistent with $S, H$; a *no* answer is consistent with $H, C$. (3) Can I walk around the gap? A *yes* answer is consistent with $S, H$; a *no* answer is consistent with $S, H, C$.

Assume that if multiple answers to a query are consistent with the true state of the gap, the distributions over such answers are uniform and independent. Note that after a few queries, the set of states consistent with all the answers is the intersection of the sets consistent with the individual answers; once this set has been reduced to one element, the robot knows the state of the gap.

Suppose the agent only has time to run one test. Then, to maximize expected utility, the robot should run test 1, because the other tests give it no chance of learning the state of the gap for certain. Now suppose that the agent has time for two tests. Then the optimal test policy is as follows: run test 2 first; if the answer is *yes*, run test 1 second; otherwise, run test 3 second. (If the true state is $S$, this is discovered with probability $\frac{1}{2}$; if it is $H$, this is discovered with probability $\frac{1}{4}$, so total expected utility is $\frac{5}{12}$. Starting with test 1 or test 3 can only give expected utility $\frac{1}{3}$.)

## 4.2 Definitions

We now define the metareasoning problem of how the agent should dynamically choose queries to ask (deliberation or information gathering actions to take) so as to disambiguate the state of the world. While the illustrative example above was for information gathering actions, the same model applies to deliberation actions for state disambiguation (such as image processing, auditory scene analysis, sensor fusing, etc.).

**Definition 5 (STATE-DISAMBIGUATION)** *We are given*

- *A set $\Theta = \{\theta_1, \theta_2, \ldots, \theta_r\}$ of possible world states;*[7]
- *A probability function $p$ over $\Theta$;*

---

[7] If there are two situations that are equivalent from the agent's point of view (the agent's optimal course of action is the same and the utility is the same), then we consider those situations to be one state. Note that two such situations may lead to different answers to the queries. For example, one situation may be that the gap is an indoor staircase, and another situation may be that the gap is an outdoor staircase. These situations are considered to be the same state, but will give different answers to the query "Am I inside?".

- *A utility function $u : \Theta \to \Re^{\geq 0}$ where $u(\theta_i)$ gives the utility of knowing for certain that the world is in state $\theta_i$ at the end of the metareasoning process; (not knowing the state of the world for certain always gives utility 0);*

- *A query set $Q$, where each $q \in Q$ is a list of subsets of $\Theta$. Each such subset corresponds to an answer to the query, and indicates the states that are consistent with that answer. We require that for each state, at least one of the answers is consistent with it: that is, for any $q = (a_1, a_2, \ldots, a_m)$, we have $\bigcup_{1 \leq j \leq m} a_j = \Theta$. When a query is asked, the answer is chosen (uniformly) randomly by nature from the answers to that query that are consistent with the world's true state (these drawings are independent);*

- *An integer $N$; A target value $G$.*

*We are asked whether there exists a policy for asking at most $N$ queries that gives expected utility at least $G$. (Letting $\pi(\theta_t)$ be the probability of identifying the state when it is $\theta_t$, the expected utility is given by $\sum_{1 \leq t \leq r} p(\theta_t)\pi(\theta_t)u(\theta_t)$.)* [8]

### 4.3 Results

Before presenting our $\mathcal{PSPACE}$-hardness result, we will first present a relatively straightforward $\mathcal{NP}$-hardness result for the case where for each query, only one answer is consistent with the state of the world. This situation occurs when the states are so specific as to provide enough information to answer every query. Our reduction is from SET-COVER.

**Definition 6 (SET-COVER)** *We are given a set $S$, a collection of subsets $\mathcal{T} = \{T_i \subseteq S\}$, and a positive integer $M$. We are asked whether any $M$ of these subsets cover $S$, that is, whether there is a subcollection $\mathcal{U} \subseteq \mathcal{T}$ such that $|\mathcal{U}| = M$ and $\bigcup_{T_i \in \mathcal{U}} = S$.*

**Theorem 3** *STATE-DISAMBIGUATION is $\mathcal{NP}$-hard, even when for each state-query pair there is only one consistent answer.*

**Proof**: We reduce an arbitrary SET-COVER instance to the following STATE-DISAMBIGUATION instance. Let $\Theta = S \bigcup \{b\}$. Let $p$ be uniform. Let $u(b) = 1$ and for any $s \in S$, let $u(s) = 0$. Let $Q = \{(\Theta - T_i, T_i) : T_i \in \mathcal{T}\}$. Let $M = N$ and let $G = \frac{1}{|S|+1}$. We claim the instances are equivalent.

---
[8]There are several natural generalizations of this metareasoning problem, each of which is at least as hard as the basic variant. One allows for positive utilities even if there remains some uncertainty about the state at the end of the disambiguation process. In this more general case, the utility function would have *subsets* of $\Theta$ as its domain (or perhaps even probability distributions over such subsets). In general, specifying such utility functions would require space exponential in the number of states, so some restriction of the utility function is likely to be necessary; nevertheless, there are utility functions specifiable in polynomial space that are more general than the one given here. Another generalization is to allow for different distributions for the query answers given. One could also attribute different execution costs to different queries. Finally, it is possible to drop the assumption that queries completely rule out certain states, and rather take a probabilistic approach.

---

First suppose there is a solution to the SET-COVER instance, that is, a subcollection $\mathcal{U} \subseteq \mathcal{T}$ such that $|\mathcal{U}| = M$ and $\bigcup_{T_i \in \mathcal{U}} = S$. Then our policy for the STATE-DISAMBIGUATION instance is simply to ask the queries corresponding to the elements of $U$, in whichever order and unconditionally on the answers of the query. If the true state is in $S$, we will get utility 0 regardless. If the true state is $b$, each query will eliminate the elements of the corresponding $T_i$ from consideration. Since $\mathcal{U}$ is a set cover, it follows that after all the queries have been asked, all elements of S have been eliminated, and we know that the true state of the world is $b$, to get utility 1. So the expected utility is $\frac{1}{|S|+1}$, so there is a solution to the STATE-DISAMBIGUATION instance.

On the other hand, suppose there is a solution to the STATE-DISAMBIGUATION instance, that is, a policy for asking at most $N$ queries that gives expected utility at least $G$. Because given the true state of the world, there is only one answer consistent with it for each query, it follows that the queries that will be asked, and the answers given, follow deterministically from the true state of the world. Since we cannot derive any utility from cases where the true state of the world is not $b$, it follows that when it is $b$, we must be able to conclude that this is so in order to get positive expected utility. Consider the queries that the policy will ask in this latter case. Each of these queries will eliminate precisely the corresponding $T_i$. Since at the end of the deliberation, all the elements of $S$ must have been eliminated, it follows that these $T_i$ in fact cover $S$. Hence, if we let $\mathcal{U}$ be the collection of these $T_i$, this is a solution to the SET-COVER instance. ∎

We are now ready to present our $\mathcal{PSPACE}$-hardness result. The reduction is from stochastic satisfiability, which is $\mathcal{PSPACE}$-complete [11].

**Definition 7 (STOCHASTIC-SAT (SSAT))** *We are given a Boolean formula in conjunctive normal form (with a set of clauses $C$ over variables $x_1, x_2, \ldots, x_n, y_1, y_2, \ldots, y_n$). We play the following game with nature: we pick a value for $x_1$, subsequently nature (randomly) picks a value for $y_1$, whereupon we pick a value for $x_2$, after which nature picks a value for $y_2$, etc., until all variables have a value. We are asked whether there is a policy (contingency plan) for playing this game such that the probability of the formula being eventually satisfied is at least $\frac{1}{2}$.*

Now we can present our $\mathcal{PSPACE}$-hardness result.

**Theorem 4** *STATE-DISAMBIGUATION is $\mathcal{PSPACE}$-hard.*

**Proof**: Let $\Theta = C \bigcup \{b\} \bigcup V$ where $V$ consists of the elements of an upper triangular matrix, that is, $V = \{v_{11}, v_{12}, \ldots, v_{1n}, v_{22}, v_{23}, \ldots, v_{2n}, v_{33}, \ldots \ldots, v_{nn}\}$. $p$ is uniform over this set. $u$ is defined as follows: $u(c) = 0$ for all $c \in C$, $u(b) = 1$, and $u(v_{ij}) = H = 2 \prod_{q \in Q} N_{ans}(q)$ for all $v_{ij} \in V$, where $N_{ans}(q)$ is the number of possible answers to $q$. The queries are as follows. For every $v_{ij} \in V$, there is a query $q_{ij} = (\{v_{ij}\}, \Theta - \{v_{ij}\})$. Additionally, for each variable $x_i$ there are the following two queries: letting $V_i = \{v_{ij} : j \geq i\}$ (that is, row $i$ in the matrix), and letting $C_z = \{c \in C : z \in c\}$, we have

- $q_{x_i} = (V_i, C, \Theta - V_i - C_{x_i} - C_{y_i}, \Theta - V_i - C_{x_i} - C_{-y_i})$;

- $q_{-x_i} = (V_i, C, \Theta - V_i - C_{-x_i} - C_{y_i}, \Theta - V_i - C_{-x_i} - C_{-y_i})$.

We have $n$ steps of deliberation. Finally, the goal is $G = \frac{2|V|H+1}{2|\Theta|}$. First suppose there is a solution to the SSAT instance, that is, there exists a contingency plan for setting the $x_i$ such that the probability that the formula is eventually satisfied is at least $\frac{1}{2}$. Now, if we ask query $q_{x_i}$ ($q_{-x_i}$), we say this *corresponds* to us selecting $x_i$ ($-x_i$); if the answer to query $q_{z_i}$ is $\Theta - V_i - C_{z_i} - C_{y_i}$ ($\Theta - V_i - C_{z_i} - C_{y_i}$), we say this corresponds to nature selecting $y_i$ ($-y_i$). Then, consider the following contingency plan for asking queries:

- Start by asking the query corresponding to how the first variable is set in the SSAT instance (that is, $q_{x_1}$ if $x_1$ is set to *true*, $q_{-x_1}$ if $x_1$ is set to *false*);
- So long as all the queries and answers correspond to variables being selected, we follow the SSAT contingency plan; that is, whenever we have to ask a query, we ask the query that corresponds to the variable that would be selected in the SSAT contingency plan if variables so far had been selected in a manner corresponding to the queries and answers we have seen;
- If, on the other hand, we get $V_i$ as an answer, we proceed to ask $v_{ii}, v_{i(i+1)}, \ldots, v_{i(n-1)}$ in that order;
- Finally, if we get $C$ as an answer, we simply stop.

We make two observations about this policy. First, if the true state of the world is one of the $v_{ij}$, we will certainly discover this. (Upon asking query $i$, which is $q_{x_i}$ or $-q_{x_i}$, we will receive answer $V_i$ and switch to $q_{ik}$ queries; then if $j < n$, query $j+1$ will be $q_{ij}$, we will receive answer $\{v_{ij}\}$, and know the state; whereas if $j = n$, we will eliminate all the other elements of $V_{ij}$ with queries $i+1$ through $n$, and know the state.) Second, if the true state is $b$, for any $i$ ($1 \leq i \leq n$), query $i$ will be either $q_{x_i}$ or $-q_{x_i}$. This will certainly eliminate all the $v_{ij}$, so we will know the state at the end if and only if we also manage to eliminate all the clauses. But now notice that each query-answer pair eliminates exactly the same clauses as the corresponding variable selections satisfy. It follows that we will know the state in the end if and only if these corresponding variable selections satisfy all the literals. But the process by which the queries and answers are selected is exactly the same as in the SSAT instance with the solution policy. It follows we discover the true state with probability at least $\frac{1}{2}$. Hence, our total expected utility is at least $\frac{|V|}{|\Theta|}H + \frac{1}{|\Theta|}\frac{1}{2} = G$. So there is a policy that achieves the goal.

Now suppose there is a policy that achieves the goal. We first claim that such a policy will always discover the true state if it is one of the $v_{ij}$. For if a policy does not manage this, then there is some $v_{ij}$ such that for some combination of answers consistent with $v_{ij}$, the policy will not discover the state. Suppose this is indeed the true state. Since each consistent answer to query $q$ occurs with probability at least $\frac{1}{N_{ans}(q)}$, it follows that the unfavorable combination of answers occurs with probability at least $\prod_{q \in Q} \frac{1}{N_{ans}(q)}$. It follows that even if we discover the true state in every other scenario, our expected utility is at most $(\frac{|V|}{|\Theta|} - \frac{1}{|\Theta|} \prod_{q \in Q} \frac{1}{N_{ans}(q)})H + \frac{1}{|\Theta|} = G + \frac{1}{|\Theta|}(-2 + \frac{1}{2}) < G$. Now, it is straightfoward to show that this implies that so long as no answer has been one of the $V_i$ or $C$, query $i$ ($1 \leq i < n$) is either $q_{x_i}$ or $-q_{x_i}$. Query $n$ may still be $q_{nn}$ under these conditions, but since queries $q_{x_n}$ and $q_{-x_n}$ are both more informative than $q_{nn}$, we may assume that the policy that achieves the target value asks one of the former two in this case as well. It follows that the part of this policy that handles the cases where no answers have been either one of the $V_i$ or $C$ corresponds exactly to a valid SSAT policy, according to the correspondence between queries/answers and variable selections outlined earlier in the proof. But now we observe, as before, that if the true state is $b$, the probability that we discover this with the STATE-DISAMBIGUATION policy is precisely the probability that this SSAT policy satisfies all the clauses. This probability must be at least $\frac{1}{2}$ in order for the STATE-DISAMBIGUATION policy to reach the target expected utility value. So there is a solution to the SSAT instance. ∎

The following theorem allows us to make any hardness result on STATE-DISAMBIGUATION go through even when restricting ourselves to a uniform prior over states, or to a constant utility function over the states.

**Theorem 5** *Every STATE-DISAMBIGUATION instance is equivalent to another STATE-DISAMBIGUATION instance with a uniform prior $p$, and to another with a constant utility function $u$ ($u(\theta_t) = 1$ for all $\theta_t \in \Theta$). Moreover, these equivalent instances can be constructed in linear time.*

**Proof**: The only relevance of $p$ and $u$ in STATE-DISAMBIGUATION is to the policy's expected utility $\sum_{1 \leq t \leq r} p(\theta_t)\pi(\theta_t)u(\theta_t)$. So, only the products $p(\theta_t)u(\theta_t)$ matter; adding a constant factor to them also makes no difference if we correct $G$ accordingly. Hence, any instance is equivalent to one where we replace $p$ and $u$ by $p'(\theta_t) = \frac{1}{|\Theta|}$ and $u'(\theta_t) = |\Theta|p(\theta_t)u(\theta_t)$. It is also equivalent to one where we replace $p$, $u$ and $G$ by $p''(\theta_t) = \frac{p(\theta_t)u(\theta_t)}{\sum_{1 \leq i \leq r}(p(\theta_i)u(\theta_i))}$, $u''(\theta_t) = 1$, $G'' = \frac{G}{\sum_{1 \leq i \leq r}(p(\theta_i)u(\theta_i))}$. ∎

## 5 Conclusion and future research

In most real-world settings, due to limited time or other resources, an agent cannot perform all potentially useful deliberation and information gathering actions. This leads to the metareasoning problem of selecting such actions carefully. Decision-theoretic methods for metareasoning have been studied in AI for the last 15 years, but there are few theoretical results on the complexity of metareasoning.

We derived hardness results for three metareasoning problems. In the first, the agent has to decide how to allocate its deliberation time across anytime algorithms running on different problem instances. We showed this to be $\mathcal{NP}$-complete. In the second, the agent has to (dynamically) allocate its deliberation or information gathering resources across multiple actions that it has to choose among. We showed this to be $\mathcal{NP}$-hard even when evaluating each individual action is very simple. In the third, the agent has to (dynamically) choose a limited number of deliberation or information gathering actions to disambiguate the state of the world. We showed that this is $\mathcal{NP}$-hard under a natural restriction, and $\mathcal{PSPACE}$-hard in general.

Our results have general applicability in that most metareasoning systems must somehow deal with one or more of these problems (in addition to dealing with other issues). The results are not intended as an argument against metareasoning or decision-theoretic deliberation control. However, they do show that the metareasoning policies directly suggested by decision theory are not always feasible. This leaves several interesting avenues for future research: 1) investigating the complexity of metareasoning when deliberation (and information gathering) is costly rather than limited, 2) developing optimal metareasoning algorithms that usually run fast (albeit, per our results, not always), 3) developing fast optimal metareasoning algorithms for special cases, 4) developing approximately optimal metareasoning algorithms that are always fast, and 5) developing meta-metareasoning algorithms to control the meta-reasoning, etc.